\documentclass{article}
\pdfoutput=1

\usepackage{arxiv}

\usepackage[utf8]{inputenc} % allow utf-8 input
\usepackage[T1]{fontenc}    % use 8-bit T1 fonts
\usepackage{hyperref}       % hyperlinks
\usepackage{url}            % simple URL typesetting
\usepackage{booktabs}       % professional-quality tables
\usepackage{amsfonts}       % blackboard math symbols
\usepackage{nicefrac}       % compact symbols for 1/2, etc.
\usepackage{lipsum}		% Can be removed after putting your text content
\usepackage{graphicx}
\usepackage{doi}

\usepackage{lineno,hyperref}
\usepackage{soul}
\usepackage{url}
\usepackage{amsmath}
\usepackage{amsthm}
\usepackage[linesnumbered, ruled, vlined]{algorithm2e}
\usepackage{multirow}
\usepackage{caption}
\usepackage{subcaption}
\modulolinenumbers[5]

\title{Class-Incremental Learning via Knowledge Amalgamation\thanks{Paper published at ECML PKDD 2022}}

\author{Marcus de Carvalho\\
	School of Computer Science and Engineering\\
	Nanyang Technological University\\
	50 Nanyang Ave, Singapore 639798 \\
	\texttt{marcus.decarvalho@ntu.edu.sg} \\
	\And
    Mahardhika Pratama\\
	School of Computer Science and Engineering\\
	University of South Australia\\
    Adelaide, Australia  \\
	\texttt{dhika.pratama@unisa.edu.au} \\
	\And
	Jie Zhang\\
	School of Computer Science and Engineering\\
	Nanyang Technological University\\
	50 Nanyang Ave, Singapore 639798 \\
	\texttt{zhangj@ntu.edu.sg} \\
	\And
	Yajuan Sun\\
	Singapore Institute of Manufacturing Technology\\
	2 Fusionopolis Way, Singapore 138634 \\
	\texttt{sun\textunderscore yajuan@simtech.a-star.edu.sg} \\
}

\date{}

\renewcommand{\shorttitle}{Class-Incremental Learning via Knowledge Amalgamation}

\hypersetup{
pdftitle={Class-Incremental Learning via Knowledge Amalgamation},
pdfsubject={cs},
pdfauthor={Marcus de Carvalho, Mahardhika Pratama, Jie Zhang, Yajuan Sun},
pdfkeywords={continual learning, transfer learning, knowledge distillation, knowledge amalgamation},
}

\begin{document}
\maketitle

\begin{abstract}
	Catastrophic forgetting has been a significant problem hindering the deployment of deep learning algorithms in the continual learning setting. Numerous methods have been proposed to address the catastrophic forgetting problem where an agent loses its generalization power of old tasks while learning new tasks. We put forward an alternative strategy to handle the catastrophic forgetting with knowledge amalgamation (CFA), which learns a student network from multiple heterogeneous teacher models specializing in previous tasks and can be applied to current offline methods. The knowledge amalgamation process is carried out in a single-head manner with only a selected number of memorized samples and no annotations. The teachers and students do not need to share the same network structure, allowing heterogeneous tasks to be adapted to a compact or sparse data representation. We compare our method with competitive baselines from different strategies, demonstrating our approach's advantages. Source-code: \url{github.com/Ivsucram/CFA}
\end{abstract}

% keywords can be removed

\section{Introduction}
\label{sec:intro}

Computational learning systems driven by the success of deep learning have obtained great success in several computational data mining and learning system as computer vision, natural language processing, clustering, and many more \cite{gama2010knowledge}. However, although deep models have demonstrated promising results on unvarying data, they are susceptible to {\it catastrophic forgetting} when applied to dynamic settings, i.e., new information overwrites past experiences, leading to a significant drop in performance of previous tasks.

In other words, current learning systems depend on batch setting training, where the tasks are known in advance, and the training data of all classes are accessible. When new knowledge is introduced, an entire retraining process of the network parameters is required to adapt to changes. This becomes impractical in terms of time and computational power requirements with the continual introduction of new tasks.

To overcome catastrophic forgetting, learning agents must integrate continuous new information to enrich the existing knowledge. The model must then prevent the new information from significantly obstructing the acquired knowledge by preserving all or most of it. A learning system that continuously learns about incoming new knowledge consisting of new classes is called a {\it class-incremental learning} agent.

A class-incremental solution showcases three properties:
\begin{enumerate}
    \item It should learn from a data stream that introduces different classes at different times,
    \item It should provide a multi-class inference for the learned classes at any requested time,
    \item Its computational requirements should be bounded, or grow slowly, to the number of classes learned.
\end{enumerate}

Many strategies and approaches in the continual learning field attempt to solve the catastrophic forgetting problem in the class-incremental scenario. Regularization techniques \cite{IMM} identify essential parameters for inference of previous tasks and avoid perturbing them when learning new tasks. Knowledge distillation methods have also been used \cite{LearningWithoutForgetting}, where knowledge from previous tasks and incoming tasks are jointly optimized. Inspired by work in reinforcement learning, memory replay has also been an important direction explored by researchers \cite{DGR}, where essential knowledge acquired from previous experiences is re-used for faster training, or retraining, of a learning agent. 

In this paper, we propose a \underline{c}atastrophic \underline{f}orgetting solution based on knowledge \underline{a}malgamation (CFA). Given multiple trained teacher models - each on a previous task - knowledge amalgamation aims to suppress catastrophic forgetting by learning a student model that handles all previous tasks in a single-head manner with only a selected number of memorized samples and no annotations. Furthermore, the teachers and the students do not need to share the same structure so that the student can be a compact or sparse representation of the teachers' models.

A catastrophic forgetting solution based on the knowledge amalgamation approach is helpful because it allows heterogeneous tasks to be adapted to a single-head final model. At the same time, knowledge amalgamation explores the relationship between the tasks without needing any identifier during the amalgamation process, being smoothly integrated into already existing learning pipelines. This approach can be perceived as a post-processing continual learning solution, where a teacher model is developed for each task and flexibly combined into a single compact model when inference is required. As a result, it does not need to maintain specific network architectures for each task. 

{\bf Contributions}:
\begin{itemize}
    \item A novel class-incremental learning approach via knowledge amalgamation which:
    \begin{itemize}
        \item Allows teachers and students to present different structures and tasks;
        \item Integrable into existing learning pipelines (including non-continual ones);
        % \item Source-code availability: \url{bit.ly/3Fer1Y0}
    \end{itemize}
\end{itemize}

\section{Related Work}
\label{sec:related-work}

A neural network model needs to learn a series of tasks in sequences in the continual learning setting. Thus, only data from the current task is available during training. Furthermore, classes are assumed to be clearly separated. As a result, catastrophic forgetting occurs when a new task is introduced and the model loses its generalization power of old tasks through learning.

Currently, there are three scenarios in which a continual learning experiment can be configured. {\it Task-incremental learning} is the easiest of the scenarios, as a model receives knowledge about which task needs to be processed. In this scenario, models with task-specific components are the standard, where the multi-headed output layer network represents the most common solution.

The second scenario, referred to as {\it domain-incremental learning}, does not have task identity available during inference, and models only need to solve the given task without inferring its task.

Finally, {\bf class-incremental learning}, the third scenario, requires that the models must solve each task seen so far while at the same time inferring its task. The currently proposed method falls into this scenario. Furthermore, most real-world problems of incrementally learning new classes of objects also belong to this scenario. 

Existing works to handle the continual learning problem are mainly divided into three categories:

\begin{itemize}
    \item {\bf Structure-based approach}: One reason for catastrophic forgetting is that a neural network's parameters are optimized for new tasks and no longer for previous ones. This suggests that not optimizing the entire network or expanding the internal model structure to deal with the new tasks while isolating old network parameters could attenuate catastrophic forgetting. PNN \cite{PNN} pioneered this approach by adding new components to the network and freezing old task parameters during training. Context-dependent gating (XdG) \cite{XDG} is a simple but popular approach to randomly assigning nodes to tasks. However, these approaches are limited to the {\it task-incremental learning} scenario by design, as task identity is required to select the correct task-specific components during training.
    
    \item {\bf Regularization-based approach}: When task knowledge is only available during training time, training a different part of the network for each task can still happen, but then the whole network is used through inference. Standard methods in this approach estimate the importance of the network parameters for the previously learned tasks and penalize future changes accordingly. Elastic Weight Consolidation (EWC) \cite{EWC} and its online counterpart (EWCo) \cite{omniglot} adopt the Fisher information matrix to estimate the importance of the network synapses. Synaptic intelligence (SI) \cite{SI} utilizes an accumulated gradient to quantify the significance of the network parameters.
    
    \item {\bf Memory-based approach}: This strategy replays old, or augmented, samples stored in memory when learning a new task. Learning without forgetting (LWF) \cite{LearningWithoutForgetting} uses a pseudo-data strategy where it labels the samples of the current tasks using the model trained on the previous tasks, resulting in training that mixes hard-target (likely category according to previous tasks) with soft-target (predicted probabilities within all classes). Gradient episodic memory (GEM) \cite{GEM} and Averaged GEM (A-GEM) \cite{A-GEM} successfully boost continual learning performance by the usage of exact samples stored in memory to estimate the forgetting case and to constraint the parameter updates accordingly. Gradient-based Sample Selection (GSS) \cite{GSS} focuses on optimizing the selection of samples to be replayed. Dark Experience Replay (DER/DER++) \cite{DER} and Function Distance Regularization (FDR) \cite{FDR} use past samples and soft outputs to align past and current outputs. Hindsight Anchor Learning (HAL) \cite{HAL} adds additional objectives into replaying, aiming to reduce forgetting of key learned data points. %iCaRL \cite{iCaRL} is a combination of replay and distillation.
    
    Alternatively, methods can also take advantage of generative models for pseudo-rehearsal. For example, Deep Generative Replay (DGR) \cite{DGR} utilizes a separated generative model sequentially trained on all tasks to generate samples from their data distribution. Additionally, knowledge distillation can be combined with DGR (DGR+distill) \cite{vandeVen2019ThreeSF} to pair generated samples with soft target knowledge.
\end{itemize}

The proposed CFA is a memory-based approach that presents a novel way to perform continual learning using knowledge amalgamation, a derivation of knowledge distillation, and domain adaptation to merge several teacher models into a single student model.

\subsection{Domain Adaptation}
\label{subsec:domain-adaptation}

Transfer learning (TL) \cite{TLsurvey} is defined by the reuse of a model developed for a task to improve the learning of another task. Neural networks have been applied to TL because of their power in representing high-level features.

While there are many sub-topics of TL, we are deeply interested in domain adaptation (DA) \cite{ben2010theory}. While there are many approaches to measure and reduce the disparity between the distributions of these two domains, Maximum Mean Discrepancy (MMD) \cite{gretton12a} and Kullback-Leibler divergence (KL) \cite{KL-divergence} are widely used in the literature. Our approach uses KL to approximate the representation of learning distributions between the teachers and the student, differing from the original knowledge amalgamation method \cite{KA}, where the MMD approach is applied.

\subsection{Knowledge Distillation}
\label{subsec:knowledge-distillation}

Knowledge distillation (KD) \cite{KnowledgeDistillation} is a method of transferring learning from one model to the other, usually by compression, where a larger teacher model supervises the training of a smaller student model. One of the benefits of KD is that it can handle heterogeneous structures, i.e., the teacher and the student do not need to share the same network structure. Instead, the teacher supervises the student training via its logits, also called the soft target. In other words, KD minimizes the distance between the student network output $\hat{z}$ and the logits $z$ from a teacher network, generated from an arbitrary input sample:

\begin{equation}
    \mathcal{L}_{KD} = ||\hat{z} - z||_2^2
\end{equation}

Although KD has become a field itself in the machine learning community, many approaches are still performed under a single teacher-student relationship, with a sharing task \cite{KnowledgeDistillation}. Contrary to these constraints, our method can process multiple and heterogeneous teachers, condensing their knowledge into a single student model covering all tasks. 

\subsection{Knowledge Amalgamation}
\label{subsec:knowledge-amalgamation}

Knowledge Amalgamation (KA) \cite{KA,KnowledgeAmalgamation} aims to acquire a compact student model capable of handling the comprehensive joint objective of multiple teacher models, each specialized in their task. Our approach extends the concept of knowledge amalgamation in \cite{Shen2019AmalgamatingKT,KA} to the continual learning environment.

\section{Problem Formulation}
\label{sec:problem-formulation}

In continual learning, within the class-incremental learning scenario, we experience a stream of data tuples $(x_i, y_i)$ that satisfies $(x_i, y_i) \stackrel{iid}{\sim} P_{t_i}(X, Y)$, containing an input $x_i$ and a target $y_i$ organized into sequential tasks $t_i \in \mathcal{T} = {1,...T}$, where the total number of tasks $T$ is unknown a priori. The goal is to learn a predictor $f: \mathcal{X} \times \mathcal{T} \rightarrow \mathcal{Y}$, which can be queried {\it at any time} to predict the target vector {\it y} associated to a test sample $x$, where $(x, y) \sim P_t$. Such test pair can belong to a task that we have observed in the past or the current task.

We define the knowledge amalgamation task as follows. Assume that we are given N teacher models ${t}_{i=1}^N$ trained a priori, each of which implements a specific task $T$. Let $\mathcal{D}_i$ denote the set of classes handled by model $t_i$. Without loss of generality, we assume $\mathcal{D}_i \neq \mathcal{D}_j, \forall i \neq j$. In other words, for any pair of models $t_i$ and $t_j$, we assume they classify different tasks. The goal of knowledge amalgamation is to derive a compact single-head student model that can infer all tasks, in other words, to be able to simultaneously classify all the classes in $\mathcal{D} = \cup_{i=1}^N \mathcal{D}_i$. In other words, the knowledge amalgamation mechanism is done in the post-processing manner where all teacher models trained to a specific task are combined into a single model to perform comprehensive classification as per its teacher models. This approach provides flexibility over existing continual learning approaches because a teacher model can be independently built for a specific task. Their knowledge can later be amalgamated into a student model without loss of generalization power.

\section{Proposed method}
\label{sec:cfa}

\begin{algorithm}[t]
\caption{CFA}
\label{alg:algorithm}
\KwIn{Teacher models $\mathcal{T}_N$, Task Memory $\mathcal{M}$, Student model $\mathcal{S}$, number of epochs}
\KwOut{Amalgamated Model $\mathcal{S}$}

\For{epoch in epochs}{
    M $\leftarrow$ shuffle(M); \# Optional \\ 
    \For{sample $m$ in $\mathcal{M}$}{
        \Begin(Joint Representation Learning:){ 
            $\mathcal{L}_M = \mathcal{L}_R = 0$; \# Loss initialization \\
            $\hat{f}_{S} \leftarrow f_{S} =\leftarrow F_{S}$; \# Student encoder\\
            \For{$T_i$ in $\mathcal{T}_N$}{
                $\hat{f}_{T_i} \leftarrow f_{T_i} \leftarrow F_{T_i}$; \# Teacher encoder \\
                $F'_{T_i} \leftarrow f_{T_i} \leftarrow \hat{f}_{T_i}$; \# Teacher decoder \\
                $\mathcal{L}_M$=$\mathcal{L}_M + H(\hat{f}_S, \hat{f}_{T_i}) - H(\hat{f}_S)$; \# Eq. \ref{eq:lm} \\
                $\mathcal{L}_R$ = $\mathcal{L}_R$ + $||F'_{T_i}$ - $F_{T_i}||_2^2$; \# Eq. \ref{eq:lr} \\
            }
        }
        
        \Begin(Soft Domain Adaptation:){
            $y_T \leftarrow \mathcal{T}_N(m)$; \# Stacked teachers’ soft output \\
            $D_{{\text{KL}_{\text{soft}}}} = H(\hat{y}_S, y_T) - H(\hat{y}_S)$; \# Eq. \ref{eq:soft_kl} \\
        }
        
        $\mathcal{L} = \alpha D_{{\text{KL}_{\text{soft}}}} + (1 - \alpha)(\mathcal{L}_M + \mathcal{L}_R)$; \# Eq. \ref{eq:final_loss} \\
        $S_{\theta} = S_{\theta} - \lambda \nabla \mathcal{L}$; \# Parameter learning
    }
}

\end{algorithm}

In this section, we introduce the proposed CFA and its details. The knowledge amalgamation element is an extension of \cite{KA,KnowledgeAmalgamation} and consists of two parts: a joint representational learning and a soft domain adaptation.

\begin{figure}
    \centering
    \includegraphics[width=0.85\linewidth]{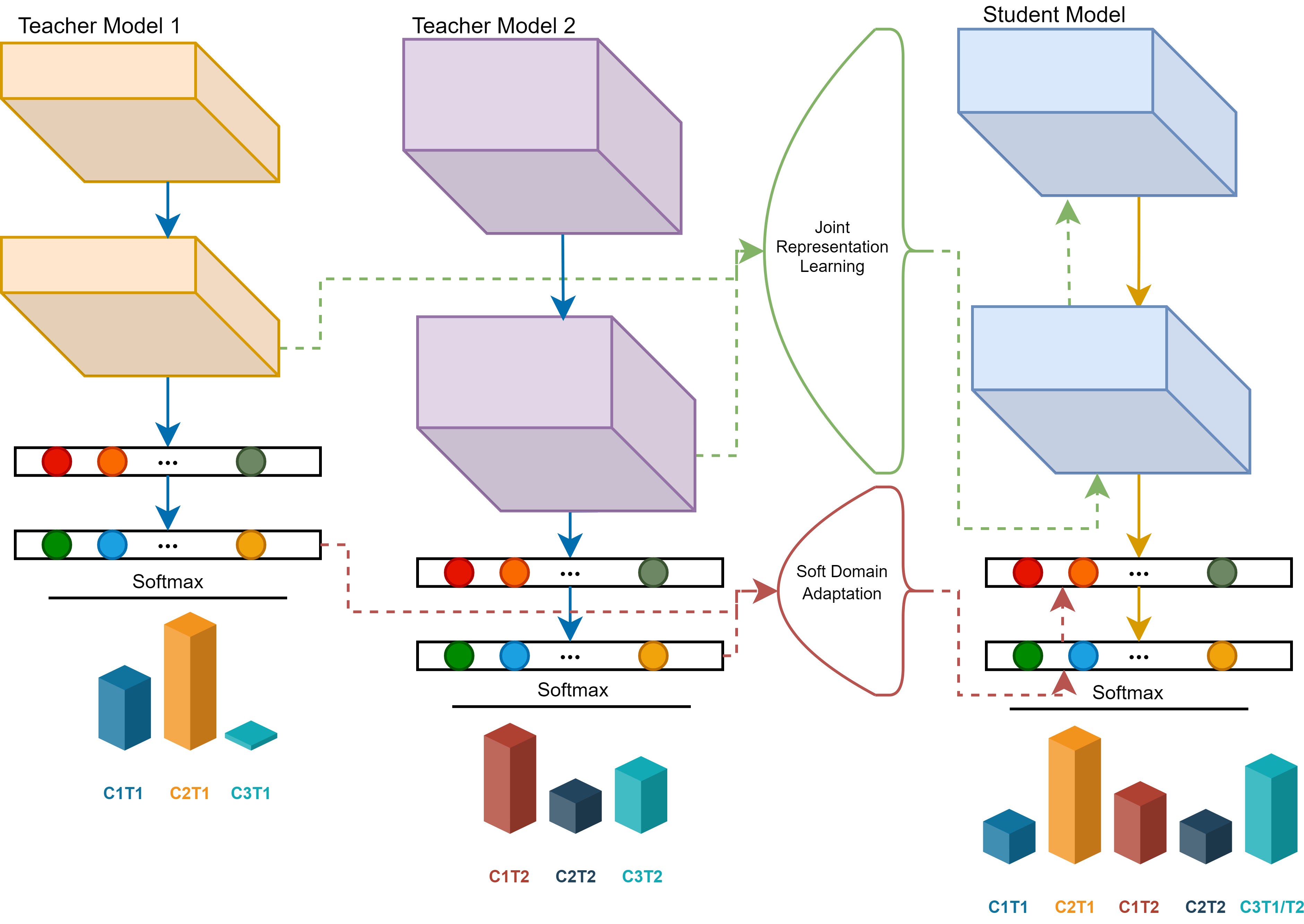}
    \caption{A summarized workflow of the proposed CFA. It consists of two parts: Joint representation learning and Soft domain adaptation. In joint representation learning, the features of the teachers (showing two here) and those to be learned by the students are first transformed into a joint space. Later on, soft domain adaptation enforces a domain invariant feature space between the student and teachers via KL.}
    \label{fig:cfa}
\end{figure}

\begin{figure}
    \centering
    \includegraphics[width=0.85\linewidth]{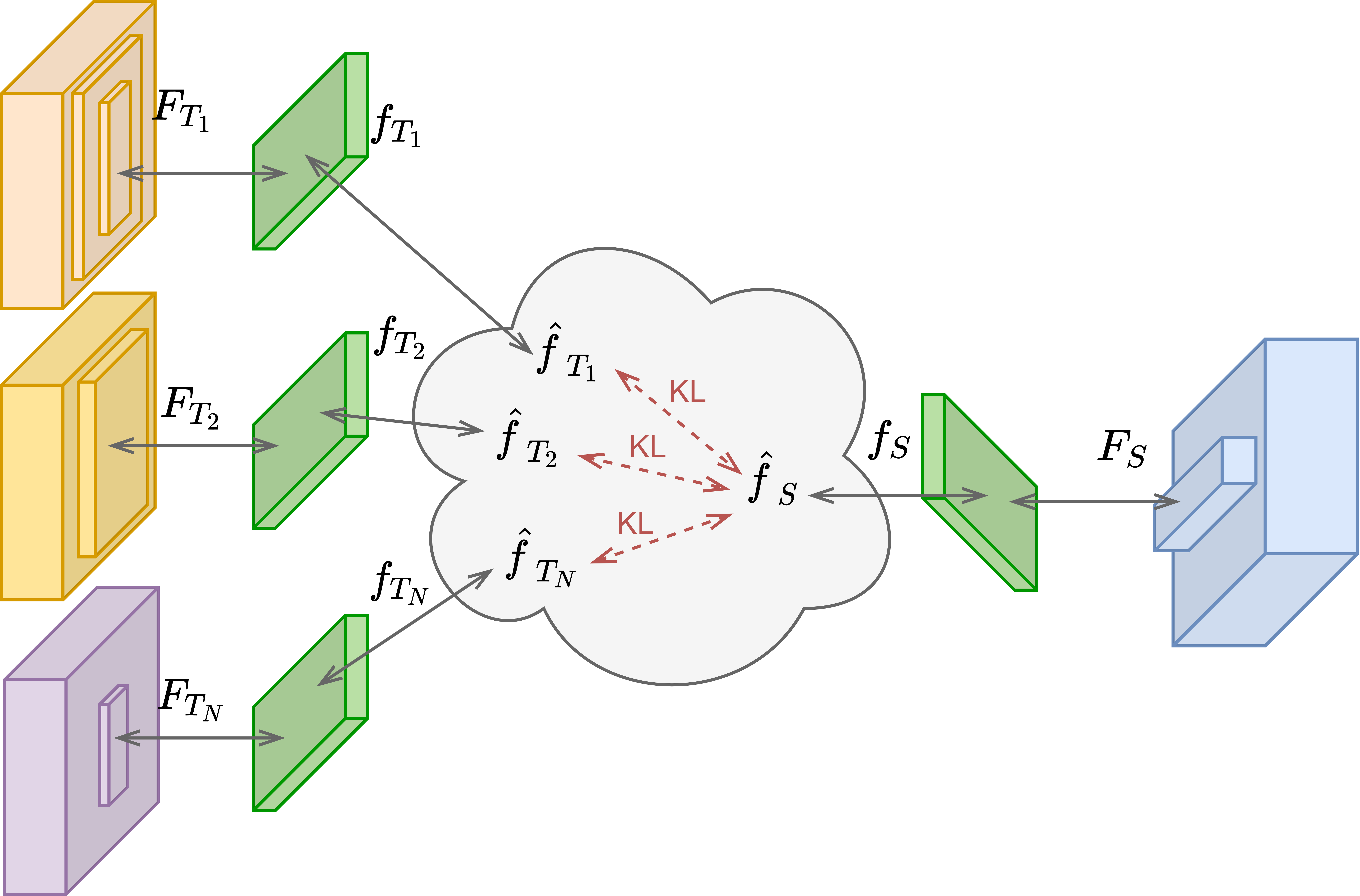}
    \caption{An illustration of the shared extractor sub-network. A sub-network is shared between the teachers and the student during the joint representation learning procedure. This shared extractor aims to create a domain-invariant space via KL, which is then decompressed back into the student model.}
    \label{fig:cfa-joint}
\end{figure}

\subsection{Joint Representation Learning}
\label{subsec:joint-representation-learning}

Figure \ref{fig:cfa} depicts the joint representation learning scheme. The features of the teachers and those to be learned from the students are first transformed into a common feature space, and then two loss terms are minimized. First, a feature ensemble loss $\mathcal{L}_M$ encourages the features of the student to approximate those of the teachers in the joint space. Then a reconstruction loss $\mathcal{L}_R$ ensures the transformed features can be mapped back to the original space with minimum possible errors. 

\subsubsection{Adaptation Layer}
\label{subsubsec:adaptation-layer}

The adaptation layer aligns the teachers' and students' output feature dimensions via a 1 x 1 convolution kernel \cite{1x1conv} that generates a predefined output length with different input sizes. Let $F_S$ and $F_{T_i}$ be respectively the original features of the student and teacher $T_i$, and $f_S$ and $f_{T_i}$ their respective aligned features. In our implementation, $f_S$ and $f_{T_i}$ have the same size of $F_S$ and $F_{T_i}$.

\subsubsection{Shared Extractor}
\label{subsubsec:shared-extractor}

Once the aligned features are derived, a naive approach would be to directly average the features of the teachers $f_{T_i}$ as that of the student $f_S$. However, due to domain discrepancy of the training data and architectural differences of the teacher networks, the roughly aligned features may remain heterogeneous. To this end, the teachers and students share the parameters of a small learnable sub-network, illustrated in Figure \ref{fig:cfa-joint}. This shared extractor consists of three consecutive residual blocks of $1$ stride. It converts $f_{T_i}$ and $f_S$ into the common representation spaces $\hat{f}_{T_i}$ and $\hat{f}_S$. In our implementation, $\hat{f}_{T_i}$ and $\hat{f}_S$ is half the size of $f_{T_i}$ and $f_S$.

\subsubsection{Knowledge Amalgamation}
\label{subsubsec:knowledge-amalgamation}

To amalgamate knowledge from heterogeneous teachers, we enforce a domain invariant feature space between the student and teachers via the KL divergence, computed as follows:

\begin{equation}
    \label{eq:dkli}
    D_{KL_i}(\hat{f}_S || \hat{f}_{T_i}) = H(\hat{f}_S, \hat{f}_{T_i}) - H(\hat{f}_S),
\end{equation}

\noindent where $H(\hat{f}_S, \hat{f}_{T_i})$ is the cross entropy of $\hat{f}_{T_i}$ and $\hat{f}_S$ and $H(\hat{f}_S)$ is the entropy of $\hat{f}_S$.

We then aggregate all such pairwise KL losses between each teacher and the student, as shown in Figure \ref{fig:cfa-joint}, and write the overall discrepancy $\mathcal{L}_M$ in the shared space as:

\begin{equation}
    \label{eq:lm}
    \mathcal{L}_M = \sum_{i=1}^N D_{KL_i},
\end{equation}

To further enhance the joint representation learning, we add an autoencoder \cite{Autoencoder} reconstruction loss between the original teachers' feature space. Let ${F'_{T_i}}$ denote the reconstructed feature of teacher $T_i$, the reconstruction loss $\mathcal{L}_R$ is defined as

\begin{equation}
    \label{eq:lr}
    \mathcal{L}_R = \sum_{i=1}^N || F'_{T_i} - F_{T_i}||_2,
\end{equation}

\subsection{Soft Domain Adaptation}
\label{subsec:soft-target-distillation}
Apart from learning the teacher's features, the student is also expected to produce identical or similar inferences as the teachers do. We thus also take the teachers' predictions by feeding unlabelled input samples to them and then supervise the student's training.

We assume that all teacher models handle non-overlapping classes, then directly stack their score vectors and use them as the student's target. A similar strategy can be used for teachers with overlapping classes, where the logits of repeating classes can be summed or averaged, but we do not explore it here. Instead of directly applying a cross-entropy loss between the student output and the teachers' soft output, as most knowledge distillation solutions do, we also enforce a domain invariant space into the discriminative (fully-connected) layers of the student by applying the KL between the student output and the stacked teachers' soft output.

Let $y_T$ denote the stacked teachers' soft output and $\hat{y}_S$ denote the corresponding student soft output, then KL is applied as:

\begin{equation}
    \label{eq:soft_kl}
    D_{KL_{soft}}(\hat{y}_S || y_T) = H(\hat{y}_S, y_T) - H(\hat{y}_S)
\end{equation}

\subsection{Final Loss}
\label{subsec:final-loss}

We incorporate the loss terms in Eqs \ref{eq:lm}, \ref{eq:lr} and \ref{eq:soft_kl} into our final loss function. The whole framework is trained end-to-end by optimizing the following objective:

\begin{equation}
    \label{eq:final_loss}
    \mathcal{L} = \alpha D_{KL_{soft}}(\hat{y}_S || y_T) + (1 - \alpha)(\mathcal{L}_M + \mathcal{L}_R)
\end{equation}

\noindent where $\alpha \in [0,1]$ is a hyper-parameter to balance the three terms of the loss function. By optimizing this loss function, the student network is trained from the amalgamation of its teachers without annotations.

\section{Experiments}
\label{sec:experiments}

We evaluate CFA and its baselines under four benchmarks. Then, an ablation study gives further insight regarding CFA memory usage and internal procedures. Finally, we executed all CFA experiments using the same structure for the teachers and students; a ResNet18 backbone \cite{Resnet} as a feature extractor and two fully-connected layers ahead of it.

\subsection{Replay Memory}
\label{subsec:replay-memory}

To retrieve {\it proper}\footnote{Meaning, related to the original data distribution} logits from the teachers, CFA uses the replay memory strategy, where it records some previous samples to be replayed during the amalgamation process. The nearest-mean-of-exemplars strategy was used to build the replay memory, but any other sample selection strategy can be used.

\subsubsection{Nearest-Mean-of-Exemplars strategy}
\label{subsubsec:nearest-mean-of-exemplars}
Consider $t_i(x)$ the logits of a teacher $t_i$ on a specific task $i$. We compute the mean exemplar for each class in class $y$ as $\mu_y = \frac{1}{||\mathcal{D}_i||}\sum_{x \in \mathcal{D}_i} t_i(x)$. A sample $x$ is then added to the memory if there is free space or by descending sorting out the memory and $x$ by their $L_2$ distance.

\subsection{Baselines setup}
\label{subsec:baselines-setup}

We set up two different configurations of CFA. CFA$_{\text{fixed}}$ uses the nearest-mean-of-exemplars replay memory strategy with a fixed memory footprint of $1000$ samples. Meanwhile, CFA$_{\text{grow}}$ uses the teachers' confidence replay memory strategy with a growing memory allowing $1000$ samples \underline{per task}. Hence, CFA$_{\text{fixed}}$ memory footprint maintains the same, independent of the number of classes learned so far\footnote{Storage of the original teacher models parameters is still required, usually in secondary memory, as HDD or SSD.}, while CFA$_{\text{grow}}$ memory footprint slowly grows are more classes are introduced.

All teachers and students have the same architecture, a pre-trained ResNet18 feature-extractor followed by two fully-connected layers, a $\mathbb{R}^{1000\times500}$ followed by a $\mathbb{R}^{500\times\text{output}}$. CFA is optimized under Adam with learning rate $\lambda = 10^{-4}$, hyper-parameter $\alpha = 0.5$, and $100$ training epochs.

The other baselines are based on the source-code release by \cite{DER}. Their configuration is also detailed in the supplemental document. The ones which are memory-based contains a memory budget of $1000$ samples \underline{per task}, making them similar to CFA$_{\text{grow}}$.

All methods have been evaluated using the same computation environment, a Windows machine with an Intel Core i9-9900K 5.0 GHz with 32GB of main memory and an Nvidia GeForce 2080 Ti.

\subsection{Metrics}
\label{subsec:metrics}

The continual learning protocol is followed, where we observe three metrics: 

\begin{equation}
    \label{eq:ACC}
    \textbf{\text{Average Accuracy: }} \text{ACC} = \frac{1}{T} \sum_{i=1}^{T} R_{T,i}
\end{equation}

\begin{equation}
    \label{eq:BWT}
    \textbf{\text{Backward Transfer: }} \text{BWT} = \frac{1}{T - 1} \sum_{i=1}^{T - 1} R_{T,i} - R_{i, i}
\end{equation}

\begin{equation}
    \label{eq:FWT}
    \textbf{\text{Forward Transfer: }} \text{FWT} = \frac{1}{T - 1} \sum_{i=2}^{T} R_{i-1,i} - \bar{b}_i
\end{equation}

\noindent where $R \in \mathbf{R}^{T x T}$ is a test classification matrix, where $R_{i,j}$ represents the test accuracy in task $t_j$ after completely learn $t_i$. The details are given by \cite{GEM}.

\begin{table}[t!]
    \begin{center}
    \begin{tabular}{cc|cccc}
    \toprule
        Baseline & Metric (\%) & \multicolumn{1}{c}{\textit{SplitMNIST}} & \multicolumn{1}{c}{\textit{SplitCIFAR10}} & \multicolumn{1}{c}{\textit{SplitCIFAR100}} & \multicolumn{1}{c}{\textit{SplitTinyImageNet}} \\
    \midrule
    \midrule
    \multirow{3}{*}{EWCo \cite{EWC,omniglot}} & ACC &  19.13 $\pm$ 0.02 &  18.57 $\pm$ 2.04 &   7.91 $\pm$ 0.79 &   7.39 $\pm$ 0.03\\
                                          & BWT & -97.37 $\pm$ 0.26 & -88.51 $\pm$ 5.18 & -83.80 $\pm$ 1.56 & -71.79 $\pm$ 0.62\\
                                          & FWT & -13.36 $\pm$ 1.38 & -10.09 $\pm$ 5.64 &  -1.02 $\pm$ 0.15 &  -0.44 $\pm$ 0.16\\
    \midrule
    \midrule
    \multirow{3}{*}{LWF \cite{LearningWithoutForgetting}} & ACC &  19.20 $\pm$ 0.06 &  16.27 $\pm$ 4.55 &   9.13 $\pm$ 0.43 &   0.41 $\pm$ 0.20\\
                                                          & BWT & -96.52 $\pm$ 0.94 & -89.24 $\pm$ 9.60 & -83.34 $\pm$ 5.57 & -50.33 $\pm$ 3.25\\
                                                          & FWT & -11.92 $\pm$ 2.01 & -11.05 $\pm$ 1.88 &   0.16 $\pm$ 0.71 &  -0.25 $\pm$ 0.03\\
                                                          
    \midrule
    \midrule
    \multirow{3}{*}{ER \cite{ER}} & ACC &  23.41 $\pm$ 0.60 &  69.07 $\pm$ 3.31  &  27.41 $\pm$ 2.94 &  12.33 $\pm$ 1.23 \\
                                  & BWT & -93.83 $\pm$ 0.92 & -24.15 $\pm$ 14.17 & -66.35 $\pm$ 1.22 & -71.07 $\pm$ 2.35 \\
                                  & FWT & -8.83  $\pm$ 3.14 & -11.84 $\pm$ 0.40  &  -0.98 $\pm$ 0.08 &   -0.5 $\pm$ 0.05 \\
    \midrule
    \multirow{3}{*}{AGEM \cite{A-GEM}} & ACC &   9.19 $\pm$ 0.65  &  13.49 $\pm$   4.12 &  0.94 $\pm$ 0.32 & 1.55 $\pm$ 0.32 \\
                                       & BWT & -40.52 $\pm$ 46.02 & -47.26 $\pm$ -47.26 &  2.99 $\pm$ 5.74 & -14.93 $\pm$ 2.12 \\
                                       & FWT &  -8.97 $\pm$ 3.54  &  -6.06 $\pm$  -6.06 &  0.74 $\pm$ 1.74 & -0.55 $\pm$ 0.20 \\
    \midrule
    \multirow{3}{*}{DER \cite{DER}} & ACC &  60.85 $\pm$ 2.87 &  72.99 $\pm$ 6.43 &  \textbf{32.60 $\pm$ 9.77} & 23.62 $\pm$ 3.30\\
                                    & BWT & -42.80 $\pm$ 3.81 & -22.71 $\pm$ 5.00 & -44.64 $\pm$ 8.54 & -52.19 $\pm$ 3.66\\
                                    & FWT & -12.25 $\pm$ 2.71 &  -9.36 $\pm$ 8.93 &  -0.93 $\pm$ 0.09 &  -0.46 $\pm$ 2.12\\
    \midrule
    \multirow{3}{*}{DER++ \cite{DER}} & ACC &  72.86 $\pm$ 0.95 &  \textbf{77.86 $\pm$ 7.59} &  \textbf{38.82 $\pm$ 8.28} & 23.94 $\pm$ 2.52\\
                                      & BWT & -24.64 $\pm$ 1.21 & -16.27 $\pm$ 5.71 & -49.03 $\pm$ 7.60 & -43.82 $\pm$ 5.95\\
                                      & FWT & -12.59 $\pm$ 0.48 &  -6.26 $\pm$ 8.81 &  -0.91 $\pm$ 0.07 &  -0.26 $\pm$ 2.16\\
    \midrule
    \multirow{3}{*}{FDR \cite{FDR}} & ACC &  78.08 $\pm$ 3.41 &  48.00 $\pm$ 5.36 &  \textbf{32.26 $\pm$ 5.51} &  13.30 $\pm$ 1.64\\
                                    & BWT & -21.73 $\pm$ 4.36 & -86.58 $\pm$ 4.37 & -62.87 $\pm$ 5.83 & -67.08 $\pm$ 1.69\\
                                    & FWT & -10.10 $\pm$ 1.13 & -11.41 $\pm$ 2.95 &  -0.87 $\pm$ 7.29 &  -0.67 $\pm$ 0.22\\
    \midrule
    \multirow{3}{*}{GSS \cite{GSS}} & ACC &  24.69 $\pm$ 0.80 &  43.96 $\pm$ 2.86 &  13.94 $\pm$ 0.30 &   9.60 $\pm$ 0.84\\
                                    & BWT & -91.69 $\pm$ 1.04 & -55.71 $\pm$ 2.57 & -78.21 $\pm$ 0.32 & -69.36 $\pm$ 0.28\\
                                    & FWT & -10.31 $\pm$ 1.96 & -10.56 $\pm$ 3.30 &  -0.39 $\pm$ 0.55 &  -0.53 $\pm$ 0.05\\
    \midrule
    \multirow{3}{*}{HAL \cite{HAL}} & ACC &  \textbf{88.25 $\pm$ 0.46} &  50.11 $\pm$ 1.18 &  11.00 $\pm$ 2.87 &   3.23 $\pm$ 0.11\\
                                    & BWT & -13.61 $\pm$ 0.62 & -47.01 $\pm$ 2.14 & -44.74 $\pm$ 1.84 & -32.68 $\pm$ 4.10\\
                                    & FWT &  -8.81 $\pm$ 3.28 & -11.70 $\pm$ 1.69 &  -0.97 $\pm$ 0.26 &  -0.24 $\pm$ 0.31\\
    \midrule
    \midrule
    \multirow{3}{*}{\textbf{CFA$_{\text{fixed}}$ (Ours)}} & ACC & 83.51 $\pm$ 1.35 &  74.96 $\pm$ 0.46 &  27.76 $\pm$ 2.28 &  23.44 $\pm$ 2.55 \\
                                                          & BWT & -7.95 $\pm$ 1.53 & -14.25 $\pm$ 1.76 & -16.41 $\pm$ 1.49 & -17.58 $\pm$ 1.89 \\
                                                          & FWT & 69.46 $\pm$ 9.41 &  54.28 $\pm$ 6.57 &  26.91 $\pm$ 3.17 &  20.49 $\pm$ 5.50 \\
    \midrule
    \multirow{3}{*}{\textbf{CFA$_{\text{grow}}$ (Ours)}} & ACC & \textbf{89.25 $\pm$ 3.66} & \textbf{79.40 $\pm$ 1.15} & \textbf{38.74 $\pm$ 3.26} & \textbf{32.50 $\pm$ 3.35} \\
                                                         & BWT & 69.77 $\pm$ 1.31  & 49.00 $\pm$ 2.78 & 11.49 $\pm$ 2.65 & 23.33 $\pm$ 3.45\\
                                                         & FWT & 91.77 $\pm$ 5.18  & 65.67 $\pm$ 8.22 & 21.84 $\pm$ 4.26 & 32.58 $\pm$ 5.12\\
    \bottomrule
    \end{tabular}
    \end{center}
    \caption{Numerical results over five execution runs.}
    \label{tab:results}
\end{table}

\subsection{Benchmarks}
\label{subsec:benchmarks}

{\it SplitMNIST} is a standard continual learning benchmark that adapts the entire MNIST problem \cite{lecun-mnisthandwrittendigit-2010} into five sequential tasks, with a total of 10 classes.

{\it SplitCIFAR10} features the incremental class problem where the full CIFAR10 problem \cite{cifar10} is divided into five sequential tasks, with a total of 10 classes.

{\it SplitCIFAR100} features the incremental class problem where the complete CIFAR100 problem \cite{cifar10} is divided into 10 sequential subsets, totalling 100 classes.

{\it SplitTinyImageNet} features the incremental class problem where 200 classes from the full ImageNet \cite{TinyImageNet} are resized to 64x64 colored pixels and divided into 10 sequential tasks.

\subsection{Numerical results}
\label{subsec:numerical-results}

We compare CFA$_{\text{fixed}}$ and CFA$_{\text{grow}}$ against one regularization-based approaches (EWCo), one knowledge distillation approaches (LWF\footnote{A multi-class implementation was put forward to deal with class-incremental learning, as in \cite{DER}}), and six memory-based approaches (AGEM, DER, DER++, FDR, GSS, HAL).

Table \ref{tab:results} presents a metric summary between the chosen baselines and benchmarks. It demonstrates that CFA$_{\text{fixed}}$ and CFA$_{\text{grow}}$ are comparable, or even stronger, in comparison with the current state-of-the-art methods, specially when we take in consideration that CFA$_{\text{fixed}}$ presents a fixed memory footprint. Furthermore, both CFA$_{\text{fixed}}$ and CFA$_{\text{grow}}$ have great BWT and FWT metrics, with CFA$_{\text{grow}}$ being the only model providing positive values to all metrics. This means that CFA signalizes some zero-shot learning \cite{GEM}, although not explicitly focused here. 

Furthermore, the most outstanding achievement of CFA is achieving good continual learning performance when applied to an offline environment while maintaining competitive results. In other words, all other methods are fully continual learning procedures, requiring an organization to shift its entire learning pipeline from scratch. In contrast, CFA leverages the power of individual teachers trained on the tasks in an online or offline environment. This scenario is expected in current organization pipelines, saving costs in an inevitable paradigm shift from offline to online learning agent technologies.

\subsection{Ablation study}
\label{subsec:ablation-study}

\subsubsection{Memory Analysis}
\label{subsubsec:memory-analysis}

\begin{table}[ht]
    \begin{center}
    \begin{tabular}{cc|ccccc}
    \toprule
        Benchmark & Memory budget & CFA$_{\text{fixed}}$ & CFA$_{\text{grow}}$ & DER & DER++ & FDR \\
    \midrule
    \midrule
    \multirow{5}{*}{\textit{Split CIFAR10}} &  100 & 48.87 $\pm$ 5.26 & \textbf{61.36 $\pm$ 2.02} & 46.77 $\pm$ 3.12 & 51.91 $\pm$ 4.21 & 39.60 $\pm$ 4.54 \\
                                            &  200 & 61.20 $\pm$ 4.35 & \textbf{69.33 $\pm$ 1.54} & 58.41 $\pm$ 3.23 & 64.92 $\pm$ 6.15 & 44.49 $\pm$ 4.31 \\
                                            &  500 & 69.53 $\pm$ 3.30 & \textbf{74.63 $\pm$ 1.20} & 65.63 $\pm$ 5.95 & \textbf{72.45 $\pm$ 6.85} & 48.20 $\pm$ 5.30 \\
                                            & 1000 & 74.96 $\pm$ 0.46 & \textbf{79.40 $\pm$ 1.15} & 72.99 $\pm$ 6.43 & \textbf{77.86 $\pm$ 7.59} & 41.91 $\pm$ 6.42 \\
                                            & 2000 & 76.45 $\pm$ 1.90 & \textbf{82.21 $\pm$ 2.52} & 73.81 $\pm$ 5.12 & 77.44 $\pm$ 8.90 & 47.39 $\pm$ 7.01 \\
    \midrule
    \midrule
    \multirow{5}{*}{\textit{Split CIFAR100}} &  100 &   7.75 $\pm$ 1.20 & \textbf{21.34 $\pm$ 2.30} & 13.23 $\pm$ 0.00 & \textbf{22.88 $\pm$ 4.90} & 12.23 $\pm$ 4.50 \\
                                             &  200 &  12.87 $\pm$ 2.12 & \textbf{26.01 $\pm$ 2.53} & 19.98 $\pm$ 0.00 & 23.78 $\pm$ 5.20 & 14.74 $\pm$ 2.24 \\
                                             &  500 &  21.23 $\pm$ 2.23 & \textbf{30.59 $\pm$ 3.54} & 26.53 $\pm$ 5.23 & \textbf{31.45 $\pm$ 6.43} & 22.26 $\pm$ 4.21 \\
                                             & 1000 &  27.76 $\pm$ 2.28 & \textbf{38.74 $\pm$ 3.26} & 32.60 $\pm$ 9.77 & \textbf{38.82 $\pm$ 8.28} & 32.26 $\pm$ 5.51 \\
                                             & 2000 &  41.50 $\pm$ 3.45 & \textbf{47.54 $\pm$ 3.18} & 36.78 $\pm$ 9.88 & 43.45 $\pm$ 8.54 & 33.12 $\pm$ 5.40 \\
    \bottomrule
    \end{tabular}
    \end{center}
    \caption{ACC(\%) metrics over different budget memories.}
    \label{tab:memory-budget-analysis}
\end{table}

Table \ref{tab:memory-budget-analysis} put the strongest baselines face to face to compare how their accuracies change over different memory budgets. CFA$_{\text{fixed}}$ maintains a competitive performance, even though it presents a fixed memory footprint. So, even though it performs similarly to DER, DER++, and FDR, it benefits from using less memory and being applied to current offline learning pipelines.

\subsubsection{Joint representation learning analysis}
\label{subsubsec:joint-representation-learning}

\begin{table}[ht]
    \begin{center}
    \begin{tabular}{c|c}
    \toprule
    \multicolumn{1}{c}{\textit{Description}} & \multicolumn{1}{c}{\textit{Split CIFAR10}} \\
    \midrule
    \midrule
    $\alpha = 1.0$ | $\text{JRL} (\times)     \text{SDA} (\checkmark)$ & 35.78 $\pm$ 10.78 \\
    $\alpha = 0.5$ | $\text{JRL} (\checkmark) \text{SDA} (\checkmark)$ & 74.96 $\pm$ 0.46 \\
    $\alpha = 0.0$ | $\text{JRL} (\checkmark) \text{SDA} (\times)$     & 69.68 $\pm$ 2.23 \\
    \bottomrule
    \end{tabular}
    \end{center}
    \caption{ACC(\%) results with varying hyper-parameter $\alpha$ of the CFA$_{\text{fixed}}$ with 1000 of memory budget, controlling the influence of the Joint Representation Learning (JRL) and Soft Domain Adaptation (SDA) into its main loss.}
    \label{tab:alpha-analysis}
\end{table}

As shown in Table \ref{tab:alpha-analysis}, Joint Representation Learning (JRL) is the main adaptation driver, responsible for driving the student's latent space to represent different tasks. Furthermore, as we are using the same architecture for the teachers and students, the difference between $\alpha = 1.0$ and $\alpha = 0.5$ is not that significant here but immensely important when dealing with entire heterogeneous structures, as noted by \cite{KA}. When JRL is disabled, the model has difficulties learning high-feature representations only with the soft domain adaptation (SDA), resulting in tremendous catastrophic forgetting.

\section{Conclusion}
\label{sec:conclusion}

This paper proposes CFA, an approach to handle catastrophic forgetting for the class-incremental environment with knowledge amalgamation. CFA can amalgamate the knowledge of multiple heterogeneous trained teacher models, each for a previous task, into a single-headed student model capable of handling all tasks altogether.

We compared CFA with a set of competitive baselines under the class-incremental learning scenario, yielding positive generalization with excellent average accuracy and knowledge transfer capabilities, backed by backward and forward knowledge transfer metrics. At the same time, CFA demonstrated some zero-shot learning aptitude and handled an enormous number of classes simultaneously.

CFA presents a novel approach towards continual learning using knowledge amalgamation, enabling easy integration to current learning pipelines and shifting from offline to online learning with a performance similar to or superior to the best of the only-online existing methods. Our approach is perceived as a post-processing approach of continual learning, distinguishing itself from existing approaches. Our future work is directed to explore the continual learning problem in multi-stream environments.

\section{Acknowledgement}
This work is financially supported by National Research Foundation, Republic of Singapore under IAFPP in the AME domain (contract no.: A19C1A0018).

\bibliographystyle{splncs04}

\end{document}